\documentclass{article}

\usepackage{microtype}
\usepackage{graphicx}
\usepackage{caption}
\usepackage{subcaption}
\usepackage{booktabs} 
\usepackage[utf8]{inputenc} 
\usepackage[T1]{fontenc}    
\usepackage{url}            
\usepackage{booktabs}       
\usepackage{amsfonts}       
\usepackage{nicefrac}       
\usepackage{microtype}      
\usepackage{xcolor}         
\usepackage{wrapfig}
\usepackage{booktabs}
\usepackage{enumitem}
\usepackage{makecell}
\usepackage{diagbox}
\usepackage{bm}
\usepackage{dsfont}
\usepackage{multirow}
\usepackage{varwidth}
\usepackage{pifont}
\usepackage{makecell}
\usepackage{wrapfig}
\usepackage{graphicx} 
\usepackage{comment}
\excludecomment{comment}
\usepackage{color}
\usepackage[colaction]{multicol}
\usepackage{colortbl}

\usepackage{algorithm}
\usepackage{xspace}
\usepackage{rotating}
\usepackage{longtable}
\usepackage{adjustbox}
\usepackage{threeparttable}
\usepackage{listings}

\usepackage{hyperref}
\definecolor{myblue}{HTML}{bfcdf0}
\definecolor{myred}{HTML}{e07f7f}

\usepackage[preprint]{icml2025}

\usepackage{amsmath}
\usepackage{amssymb}
\usepackage{mathtools}
\usepackage{amsthm}

\usepackage[capitalize,noabbrev]{cleveref}

\theoremstyle{plain}

\theoremstyle{definition}

\theoremstyle{remark}

\usepackage[textsize=tiny]{todonotes}

\icmltitlerunning{Mixture of Universal Experts: Scaling Virtual Width via Depth-Width Transformation}

\begin{document}
\newcommand{\aname}{\textsc{MoUE}\xspace}
\twocolumn[
\icmltitle{
Mixture of Universal Experts: \\ Scaling Virtual Width via Depth-Width Transformation
}



\icmlsetsymbol{equal}{*}

\begin{icmlauthorlist}
\icmlauthor{Yilong Chen}{iie,ucas,equal}
\icmlauthor{Naibin Gu}{iie,ucas,equal}
\icmlauthor{Junyuan Shang}{baidu}
\icmlauthor{Zhenyu Zhang}{baidu}
\icmlauthor{Yuchen Feng}{iie,ucas}
\icmlauthor{Jiawei Sheng}{iie,ucas}\\
\icmlauthor{Tingwen Liu}{iie,ucas}
\icmlauthor{Shuohuan Wang}{baidu}
\icmlauthor{Yu Sun}{baidu}
\icmlauthor{Hua Wu}{baidu}
\icmlauthor{Haifeng Wang}{baidu}
\end{icmlauthorlist}

\icmlaffiliation{iie}{Institute of Information Engineering, Chinese Academy of Sciences}
\icmlaffiliation{ucas}{School of Cyber Security, University of Chinese Academy of Sciences}
\icmlaffiliation{baidu}{Baidu Inc}

\icmlcorrespondingauthor{Tingwen Liu}{liutingwen@iie.ac.cn}

\icmlkeywords{Machine Learning, ICML}

\vskip 0.3in
]



\printAffiliationsAndNotice{} 

\begin{abstract}
Mixture-of-Experts (MoE) decouples model capacity from per-token computation, yet their scalability remains limited by the physical dimensions of depth and width.
To overcome this, we propose \textbf{Mixture of Universal Experts (\aname)}, a MoE generalization introducing a novel scaling dimension: \emph{Virtual Width}.
In general, MoUE aims to reuse a universal layer-agnostic expert pool across layers, converting depth into virtual width under a fixed per-token activation budget.
However, two challenges remain: 
a routing path explosion from recursive expert reuse, and a mismatch between the exposure induced by reuse and the conventional load-balancing objectives.
We address these with three core components:
a \textbf{Staggered Rotational Topology} for structured expert sharing, a \textbf{Universal Expert Load Balance} for depth-aware exposure correction, and a \textbf{Universal Router} with lightweight trajectory state for coherent multi-step routing.
Empirically, MoUE consistently outperforms matched MoE baselines by up to \textbf{1.3\%} across scaling regimes, enables progressive conversion of existing MoE checkpoints with up to \textbf{4.2\%} gains, and reveals a new scaling dimension for MoE architectures.

\end{abstract}

\section{Introduction}

Large language models, predominantly built on the Transformer architecture~\cite{vaswani2017attention}, have driven rapid progress in natural language processing. This progress can be shaped by scaling laws, where improvements are closely coupled to increases in model size and training compute~\cite{kaplan2020scaling,hoffmann2022training}. However, scaling dense Transformers incurs a structural cost: increasing parameters tends to increase both computation (FLOPs) and memory footprint proportionally. The Mixture-of-Experts (MoE) framework alleviates this tension by introducing conditional computation~\cite{fedus2022switch,deepseekai2024deepseekv2strongeconomicalefficient,feng2025dive,wang2026ernie50technicalreport}, decoupling the total parameters from the per-token activated parameters.

Despite their efficiency, standard MoE architectures remain constrained along two dimensions \cite{shazeer2017sparsemoe,lepikhin2021gshard,fedus2022switch}. In the \textit{depth} dimension, they rely on a fixed layer stacking, which makes it difficult to fully exploit depth for optimizing deep parameters \cite{shazeer2017sparsemoe,lepikhin2021gshard}; moreover, without an explicit recursive structure, they are not naturally suited to fitting complex algorithms that require reusable multi-step computation \cite{dehghani2019universal,chen2025innerthinkingtransformerleveraging}. In the \textit{width} dimension, performance is often bounded by the number of experts, yet scaling experts brings substantial system overhead and engineering cost \cite{fedus2022switch,li2023lina}. This motivates a fundamental question:

\textit{Is there an architecture that can expand model capacity by reusing a model's own depth, while introducing as little additional compute or memory overhead as possible?}

\begin{figure}[t]
\centering
\includegraphics[width=8cm]{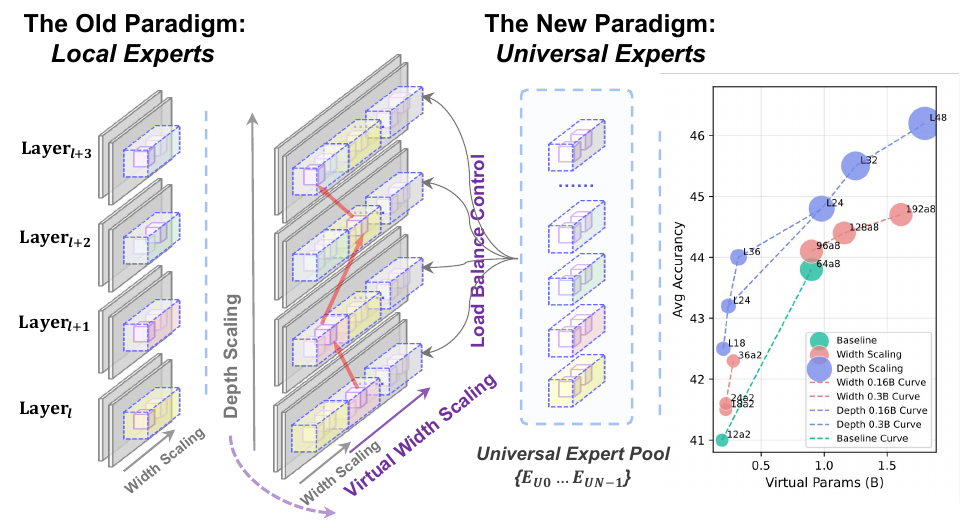} 
\caption{Overview of \aname. A shared pool of Universal Experts is accessible from multiple layers, enabling recursive reuse under a fixed activation budget.}
\label{fig:activate3d_legacy}
\vspace{-5mm}
\end{figure}

For a solution, we view \emph{depth} and \emph{width} as two physical scaling dimensions. \aname introduces a new understanding dimension for MoE, which we call \textbf{Virtual Width}: by reusing a shared pool of layer-agnostic Universal Experts (UEs) across layers, \aname turns additional depth into additional effective width through composable expert paths, while keeping the per-token activation budget fixed. Intuitively, \aname can be seen as converting a deep, narrow MoE into an extremely wide MoE layer that performs recursive computation over depth. Importantly, \aname is a strict superset of standard MoE and is seamlessly compatible: disabling cross-layer reuse recovers the conventional layer-local MoE, whereas enabling reuse decouples physical parameter storage from functional connectivity and unlocks path-composition capacity without requiring expert memory to grow in lockstep.

However, realizing this potential raises two key training challenges. First, cross-layer reuse induces a \textbf{combinatorial explosion of routing paths}, making optimization harder and less stable. Second, the standard load-balancing objective does not account for \textbf{balanced reuse across depth}, which introduces bias toward repeatedly exposed experts and can suppress the shared pool.
We address these challenges with three complementary components: 
(i) a \textbf{Staggered Rotational Topology} that groups multiple layers into shared connectivity windows, where layers within a window share the same reachable UE set while the window assignment evolves with depth; (ii) a \textbf{Universal Expert Load Balance (UELB)} strategy that explicitly introduces a depth-wise balancing dimension for experts under multi-layer reuse, combined with a warmup schedule to stabilize early exploration; and (iii) a \textbf{Universal Router} that maintains a lightweight trajectory state to make routing decisions coherent across recursive computation steps.


For evaluation, we pre-train \aname models at multiple scales and compare them with matched MoE baselines. In a width-expansion setting, training \aname from scratch achieves a 1.3\% relative performance gain over MoE without increasing parameters or compute. In a depth-expansion setting, \aname improves performance by 2.5\% while keeping FFN parameters unchanged. Across scaling experiments, \aname establishes a new frontier under both activated-parameter and total-parameter budgets. Moreover, thanks to its compatibility, \aname can be warm-started from arbitrary MoE checkpoints, with gains increasing as the universal expert pool grows; in continual pre-training, this progressive conversion yields an average 4.2\% relative improvement, and the advantage persists through large-scale supervised fine-tuning. In short, \aname reframes MoE scaling with \textbf{Virtual Width}, turning depth into reusable capacity via cross-layer expert reuse under a fixed activation budget.
\section{Preliminary}

\textbf{The Mixture-of-Experts (MoE) }architecture transforms the dense computational paradigm of Transformers by introducing conditional computation~\cite{fedus2022switch}. Formally, an MoE layer replaces the static feed-forward network (FFN) with a parameterized set of expert functions $\mathcal{E} = \{E_i\}_{i=1}^{N}$. Given an input token representation $x \in \mathbb{R}^{d}$, a gating network $G_\theta: \mathbb{R}^{d} \rightarrow \Delta^{N-1}$ computes a routing distribution $P(x) = \operatorname{softmax}(W_g x)$. To decouple model capacity from inference cost, a sparse selection mechanism $\mathcal{T}(x) = \text{TopK}(P(x), k)$ is applied, activating only a subset of experts. The layer output is the probability-weighted sum of these activated experts:
\begin{equation}
    \small
    y = \sum_{i \in \mathcal{T}(x)} \frac{P_i(x)}{\sum_{j \in \mathcal{T}(x)} P_j(x)} E_i(x).
\end{equation}
While this architecture successfully scales parameter count, it remains bound by a fixed depth topology and linear memory growth with respect to expert count, limiting its ability to model complex recursive dependencies or scale width efficiently without prohibitive resource costs.

\textbf{Standard Load Balancing Optimization.}
A critical challenge in training MoE models is the \textit{expert collapse} phenomenon. To counteract this, standard protocols optimize an auxiliary load-balancing loss $\mathcal{L}_{\text{balance}}$ alongside the task loss. This is typically formulated as the dot product of the average routing probability vector $P$ and the expert utilization fraction vector $f$ over a batch of $N$ tokens:
\begin{equation}
\small
    \mathcal{L}_{\text{balance}} = N \sum_{i=1}^{N} f_i \cdot P_i.
\end{equation}
This objective inherently assumes a uniform prior over expert utility. However, as we shall distinguish in subsequent sections, this formulation is structurally inadequate for architectures employing recursive expert reuse, where expert utility can be inherently heterogeneous by design.

\section{Motivation: Case for Parameter Reuse}

Standard practice implicitly assumes that increasing computational depth must come from introducing new, layer-specific parameters at every layer. We challenge this view and posit that deep networks exhibit substantial \textbf{functional redundancy}: certain specialized operators are repeatedly required and invoked across different levels of semantic abstraction, so strict layer-wise parameter partitioning is neither necessary nor efficient.

\begin{figure}[t]
\centering
\includegraphics[width=8cm]{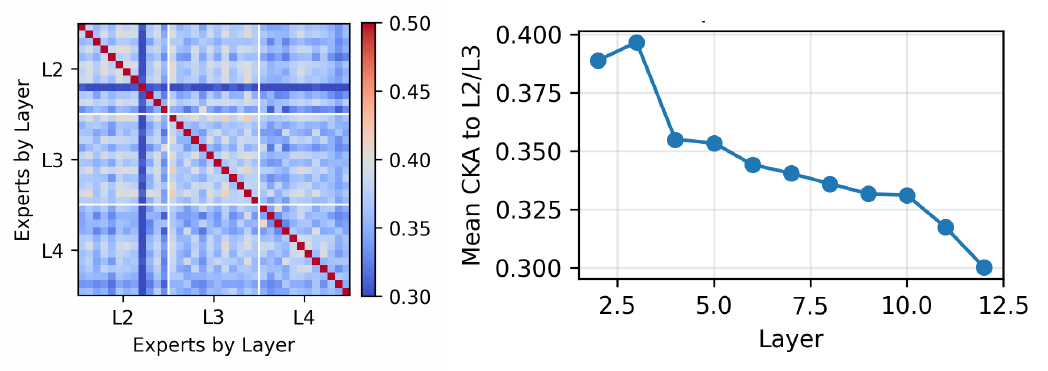}
\caption{Expert similarity heatmap across shallow (L2, L3) and deep (L8, L9) layers in a trained MoE. Adjacent layers exhibit strong similarity, and specific experts re-emerge across distant layers, motivating cross-layer expert reuse.}
\label{fig:expert_similarity}

\end{figure}

To evaluate this hypothesis, we analyze the \textbf{CKA similarity} of expert weights in a trained standard MoE~\cite{chen-etal-2024-lemon,Chen2024DHALD}. First, we observe clear \textbf{local functional persistence}: within the consecutive window of L2--L4, many expert pairs achieve CKA larger than 0.4, and the overall CKA similarity is commonly above 0.3. We further repeat the same analysis on a larger \textbf{Qwen-30B (Activated 3B)} model after SFT, where expert similarities are even higher; detailed results are provided in the appendix.

More importantly, we find pronounced \textbf{long-range similarity}. While adjacent layers are indeed much more similar, the very first layers and the very last layers still retain more than 30\% similarity. This suggests that certain functional operators are not simply replaced as depth increases, but persist (or reappear) across large depth gaps.

These findings suggest that enabling cross-layer expert reuse can mitigate the redundancy caused by repeatedly learning similar expert functions at different depths in standard MoE, thereby allowing more efficient utilization of depth and increased effective capacity through parameter sharing and path composition under a fixed activation budget.

\begin{figure*}[t]
\begin{minipage}[c]{\textwidth}
\centering
	  \includegraphics[width=0.95\textwidth]{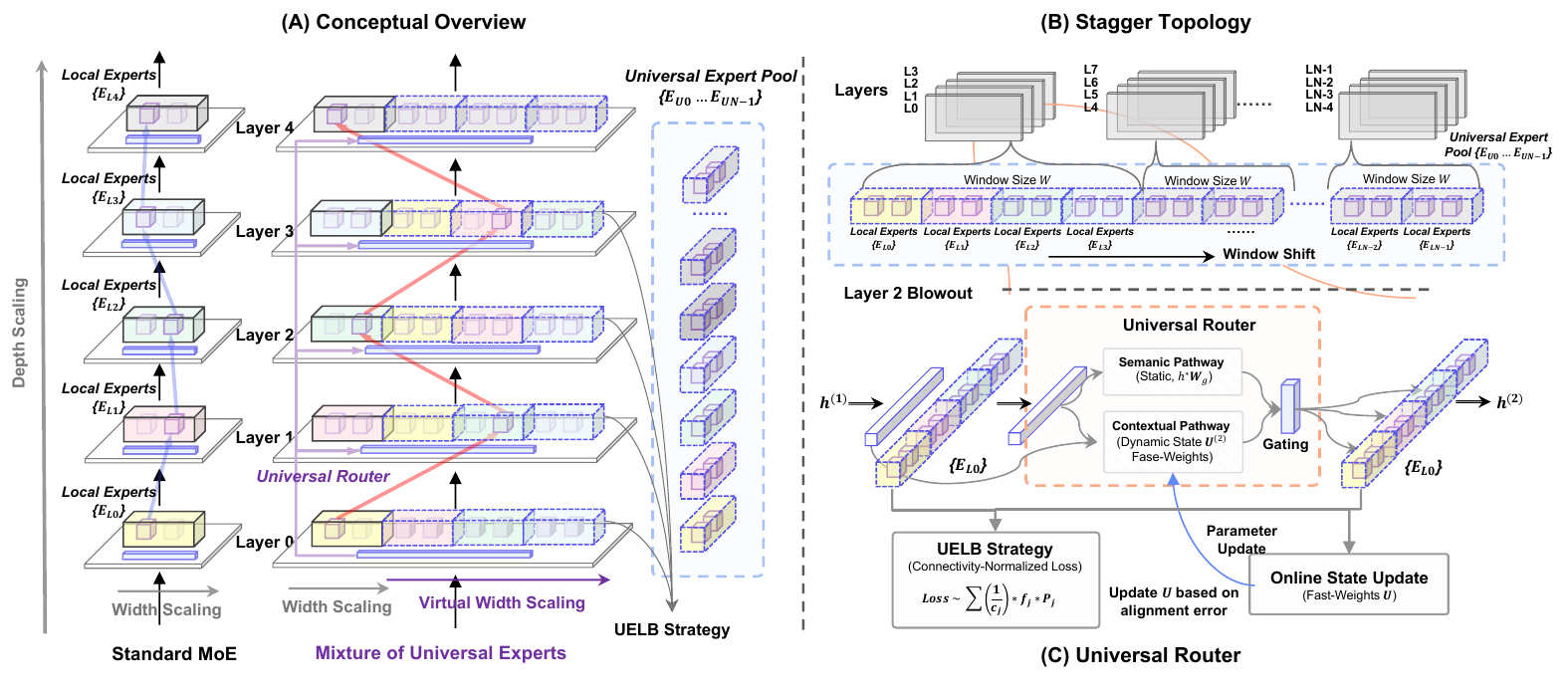} 
	  \caption{
	  An illustration of \aname. Each layer routes tokens to a small set of experts, consisting of layer-local experts and a shared pool of Universal Experts (UEs). UEs are connected to multiple layers (under a constrained topology), enabling cross-layer reuse while keeping the per-token activation budget fixed.
	  }\label{fig:main}
\end{minipage}
\end{figure*}

\section{The Mixture of Universal Experts (\aname)}

\subsection{General Framework}
\label{sec:moue.general}

To relax the rigid layer-partitioning assumption in standard MoEs, we propose \aname, which reinterprets a deep MoE not as a static stack of disjoint expert sets, but as a recursive computation over a unified addressable expert space.

\textbf{Recursive Formulation.}
Consider a Transformer of depth $L$. We decompose experts into layer-local sets $\mathcal{E}^{\text{local}}_{\ell}$ and a shared pool of Universal Experts (UEs) $\mathcal{E}^{\text{u}}$, with $\mathcal{E}^{\star}=(\bigcup_{\ell}\mathcal{E}^{\text{local}}_{\ell})\cup\mathcal{E}^{\text{u}}$. \aname is defined by a connectivity mapping $\mathcal{C}:\mathcal{E}^{\star}\times\{1,\dots,L\}\rightarrow\{0,1\}$ that specifies which experts are reachable at each layer. The MoE update at layer $\ell$ becomes a conditional query over this topology:
\begin{equation}
\begin{split}
\small
    h^{(\ell)} &= h^{(\ell-1)} + \\
    &\sum_{E_i \in \mathcal{E}^{\star}} \mathcal{C}(E_i, \ell)\cdot \pi^{(\ell)}_i(h^{(\ell-1)})\cdot E_i(h^{(\ell-1)}).
\end{split}
\end{equation}
For UEs, $\mathcal{C}(u,\ell)$ can be active for multiple layers, enabling cross-layer reuse: the same expert parameters can be invoked at different semantic stages, decoupling parameter storage from depth-wise computation.

\textbf{Theoretical Capacity and Optimization Bottleneck: Virtual Width}
Cross-layer expert reuse unlocks substantial combinatorial capacity, which we term \textbf{Virtual Width}. While standard MoE expands compositional space by increasing physical width (more experts per layer, and thus linear memory growth with depth), allowing a shared pool of $N_u$ experts at each layer enables the router to form exponentially many depth-$L$ expert paths:
\begin{equation}
\small
    |\mathcal{T}|_{\text{\aname}} \approx \left( \binom{N_u}{k} \right)^L.
\end{equation}
This exponential growth leverages depth-wise composition without increasing expert storage beyond $N_u$. However, such Virtual Width introduces optimization challenges: unconstrained reuse creates a vast routing space, weakens gradient signals, and may cause repeated selection of the same experts. Thus, effective utilization of this capacity requires structural constraints and tailored training objectives.

\subsection{Structural Optimization: Staggered Connectivity}
\label{sec:moue.staggered}

To make Virtual Width trainable, we need to constrain connectivity. In principle, there are many possible reuse topologies; in our exploration, we found that a simple yet structured design works best in practice: the \textbf{Staggered Rotational Topology}. It acts as a structural regularizer on the connectivity map $\mathcal{C}$, enforcing a controlled evolution of the reachable expert space with depth.

\textbf{Two-Level Expert Ring with Staggered Rotation.}
We organize the expert pool using a hierarchical two-level ring structure. At the coarse level, a large sliding window defines the union of experts accessible to a connectivity group comprising $G$ consecutive layers, ensuring that all layers within the group share the same set of reachable experts. Within this window, a smaller sliding window allocates a private subset of experts to each layer, while the remaining experts constitute a shared pool of Universal Experts (UEs) available for cross-layer reuse. To further enhance flexibility and diversity in expert reuse, we index all UEs on a ring and apply a staggered rotation mechanism: for each connectivity group, the shared UE window shifts by a stride $s$ along the ring, such that the set of accessible UEs for group $g(\ell)=\lfloor(\ell-1)/G\rfloor$ at layer $\ell$ is given by
\begin{equation}
\begin{split}
\small
    \mathcal{C}&(E_j, \ell) = \\ 
    &\mathbb{I}\left(j \in \left[ g(\ell) \cdot s, \dots, g(\ell) \cdot s + W - 1 \right] \pmod{N_u} \right),
\end{split}
\end{equation}
where $N_u$ is the total number of UEs and $W$ is the window size. This design enforces local specialization via layer-private experts while enabling controlled, smoothly evolving cross-layer reuse through staggered rotation of the shared expert pool. As a result, the model achieves a principled balance between specialization and reuse, supporting both efficient optimization and rich compositional capacity.

\textbf{Implementation.}
We realize $\mathcal{C}$ as per-layer allow-lists over global expert IDs, masking unreachable experts (routing logits set to $-\infty$) before Top-$k$ selection. This reduces the per-layer routing space from $N_u$ to $W$, transforming global routing into a sequence of localized decisions and improving optimization stability.\footnote{Routing complexity is reduced from $\mathcal{O}(L \cdot N_u)$ (all-to-all) to $\mathcal{O}(L \cdot W)$.} Additional design variants are discussed in Appendix~\ref{app:staggered_full}.In practice, the maximum window size can be chosen based on the expert-parallel (EP) size so that all experts within the same node are reachable. This keeps routing within a single node and avoids introducing additional inter-node communication overhead.

\subsection{Universal Expert Load Balance (UELB)}
\label{sec:moue.uelb}

\begin{figure}[t]
\centering
\includegraphics[width=8cm]{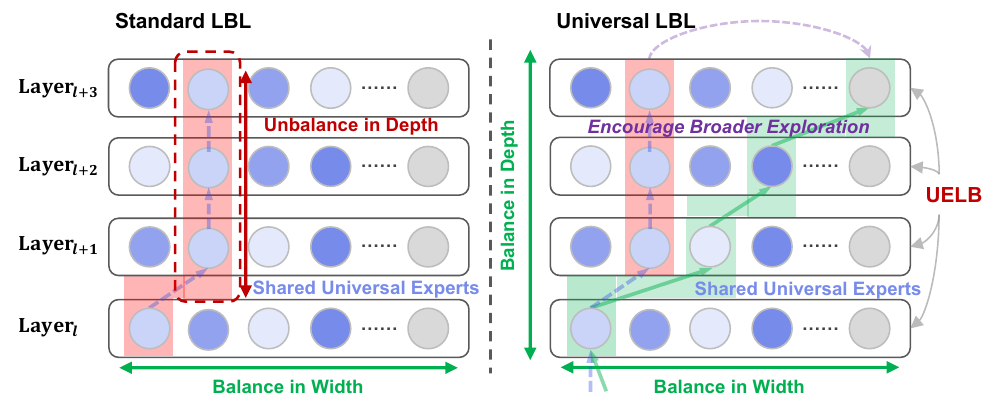}
\caption{
LBL only focuses on balancing expert selection within each layer, while ULBL promotes balanced and diverse expert selection both across depth and width.
}
\label{fig:lbl_uelb}

\end{figure}

While staggered connectivity makes routing feasible, it exposes a fundamental mismatch: standard load balancing offers weak optimization signals for recursive, multi-step expert reuse. Because \aname relies on exponentially many depth-wise expert paths, the objective must encourage not only intra-layer balance but also inter-layer diversity and path coverage. Traditional load balancing struggles due to (i) \textbf{Heterogeneous Exposure}: shared experts are over-penalized for being reachable from many layers; and (ii) \textbf{Combinatorics Blindness}: it ignores path redundancy across layers, causing underutilization of Virtual Width.

Based on these observations, we propose \textbf{Universal Expert Load Balance (UELB)}. The core innovation is to introduce a \emph{reuse-aware depth-wise} balancing dimension: ``balance'' is no longer defined purely within a layer, but is calibrated across recursive reuse opportunities over depth. The guiding principle is simple: load balancing should be enforced \emph{relative to topological opportunity}. That is, the objective should penalize an expert only when it is over-preferred \emph{within the contexts where it is available}, rather than because it is reachable more often by design.

Concretely, let $c_j=\sum_{\ell}\mathcal{C}(E_j,\ell)$ denote expert $j$'s topological degree (exposure count). UELB rescales the universal branch by $1/c_j$, so the objective reflects \emph{per-exposure} utilization rather than cross-layer aggregated usage:
\begin{equation}
\small
\begin{aligned}
\mathcal{L}_{\text{UELB}}
&= \alpha_{\text{loc}} \sum_{\ell}\sum_{i \in \mathcal{E}^{\text{loc}}}
   \bar{f}_i \bar{P}_i \\
&\quad + \alpha_{\text{u}} \sum_{j \in \mathcal{E}^{\text{u}}} \frac{1}{c_j}
   \Bigg(
      \sum_{\ell=1}^{L}
      \mathcal{C}(E_j,\ell)\,
      \bar{f}_j^{(\ell)} \bar{P}_j^{(\ell)}
   \Bigg).
\end{aligned}
\end{equation}
where $\bar{f}$ and $\bar{P}$ are the average routing frequency and expected utilization. The normalization serves two purposes: it removes architecture-induced penalties from heterogeneous exposure (decoupling reachability from preference), and it provides an optimizable depth-wise balance signal that encourages more uniform coverage of the shared expert space across recursive steps, thereby converting combinatorial path complexity into usable capacity.

In practice, we estimate UELB group-wise (layers sharing the same reachability mask) and compute a group-level auxiliary loss with a scale correction to avoid an artificial loss shrinkage due to concatenation. We further apply a lightweight warmup to encourage early exploration of the shared pool and prevent premature path collapse.

\subsection{Universal Router}
\label{sec:moue.router}

Staggered connectivity makes the routing search space local, and UELB removes exposure-induced bias while promoting depth-wise balance. Still, a standard router treats layer decisions as independent, which is at odds with \aname: useful capacity comes from coherent multi-step compositions over depth. We therefore introduce the \textbf{Universal Router}, which augments routing with a lightweight trajectory state so decisions can depend on the evolving computation path.

\textbf{Dual-Pathway Routing.}
We decompose routing logits into a semantic pathway and a contextual pathway:
\begin{equation}
    z^{(\ell)} = h^{(\ell)} W_g + b_g + \beta \cdot \text{Sim}(h^{(\ell)}, U^{(\ell)}).
\end{equation}
The semantic term performs standard affine matching. The contextual term uses a state matrix $U^{(\ell)}$ as fast weights over depth: with a projected key $k^{(\ell)}=h^{(\ell)}W_k$, we define $\text{Sim}(h^{(\ell)},U^{(\ell)})_j=\langle u_j^{(\ell)},k^{(\ell)}\rangle$, biasing selection toward experts consistent with the current trajectory.

\textbf{Online Fast-Weight Update.}
After routing, we update $U^{(\ell)}$ online as a forward-only state without backpropagation. Let $K$ be the token key matrix, $p^*$ be a detached routing target, $\hat p=\operatorname{softmax}(K U^\top)$, and $\delta=\hat p-p^*$. We apply
\begin{equation}
    U^{(\ell)} \leftarrow U^{(\ell)} - \eta_\ell \cdot \frac{\delta^\top K}{N_{\text{tok}}}.
\end{equation}
This keeps activation memory unchanged relative to a standard MoE and adds only a small per-layer overhead compared to expert FFNs. Full description in Appendix~\ref{app:router_full}.

\subsection{Progressive Warm-Start}

Training \aname from scratch is the cleanest setting but can be costly. We therefore propose a \textbf{Progressive Transformation Strategy} to convert an existing pre-trained MoE checkpoint into \aname with a curriculum-style transition.

\textbf{1. Initializing the Universal Pool.}
We construct the initial UE pool $\mathcal{E}^{\text{u}}$ by cloning a representative subset of experts from the source MoE. Rather than expensive clustering, we use a simple heuristic: experts with high activation rates in intermediate layers, which tend to be more general-purpose, are selected as UEs; the remaining experts stay layer-local to preserve layer-specific knowledge.

\textbf{2. Curriculum Routing Warmup via Logit Suppression.}
To introduce cross-layer reuse without disrupting the pre-trained feature space, we keep the original local router and add a parallel universal projection. We then apply a time-dependent negative bias to the universal branch:
\begin{equation}
    z^{(\ell)}_i =
    \begin{cases}
        h^{(\ell-1)} \cdot W_{r}^{\text{local}}[\cdot, i] & \text{if } E_i \in \mathcal{E}^{\text{local}}_{\ell} \\
        h^{(\ell-1)} \cdot W_{r}^{\text{u}}[\cdot, i] - \beta(t) & \text{if } E_i \in \mathcal{E}^{\text{u}}
    \end{cases}
\end{equation}
with $\beta(t)\ge 0$ annealed from a large value (suppressing UE usage at the start) to $0$, so the model gradually shifts from the original layer-local behavior to topology-aware reuse.
Full details in Appendix~\ref{app:warmstart_full}.

\section{Experiments}

\subsection{Experimental Setup}\label{sec:exp.setup}

\textbf{Data and Training.} We pre-train on samples from OLMo 2 Mix 1124 (3.4T tokens) and follow a fixed data-to-parameter ratio, reserving 2M tokens for validation. All models are trained with Megatron on H800 GPUs using sequence length 4096; we match tokens and activated budgets for iso-parameter and iso-compute comparisons. For warm-start conversion, we initialize from MoE checkpoints and perform continued training under matched settings.

\textbf{Baselines and Scaling Settings.} We evaluate with lm-evaluation-harness~\cite{eval-harness} against two Qwen-3 style MoE backbones spanning different scales and routing granularities: (i) MoE-160M with 12 experts and top-$2$ routing, and (ii) MoE-700M with 64 experts and top-$8$ routing. Our main comparisons study two settings. \textbf{Width expansion:} \aname increases capacity only through Virtual Width (larger path space), without increasing activated parameters or total physical parameters. \textbf{Depth expansion:} \aname performs depth scaling by sharing FFN parameters across depth. We report three budgets: activated parameters (Act), total physical parameters (TP), and virtual parameters (VP), where VP counts the effective capacity including Virtual Width. Full details are in Appendix~\ref{app:exp_setup_full}.

\begin{table*}[t]
\centering
\caption{Main results of MoUE width/depth expansion on two Qwen-3 style MoE baselines (MoE-0.16B and MoE-0.7B). We expand the \textbf{virtual width} ($\times2,\times3,\times4$) and \textbf{depth} ($\times2,\times3$) via cross-layer expert reuse.}
\vskip 0.15in
\setlength\tabcolsep{5pt}
\footnotesize
\begin{tabular}{@{}l c c c cccccc cc c c @{}}
\toprule
\small
\multirow{2}{*}{\raisebox{-0.5\height}{\textbf{Model}}}
& \multirow{2}{*}{\raisebox{-0.5\height}{\textbf{Act.}}}
& \multirow{2}{*}{\raisebox{-0.5\height}{\textbf{TP.}}}
& \multirow{2}{*}{\raisebox{-0.5\height}{\textbf{VP.}}}
& \multicolumn{6}{c}{\scriptsize\textbf{Commonsense \& Reading Comprehension}}
& \multicolumn{2}{c}{\scriptsize\textbf{Continued}}
& \multicolumn{1}{c}{\scriptsize\textbf{LM}}
& \multirow{2}{*}{\raisebox{-0.5\height}{\textbf{Avg.}}} \\
\cmidrule(lr){5-10} \cmidrule(lr){11-12} \cmidrule(lr){13-13}
& & & & \scriptsize\textbf{SciQ} & \scriptsize\textbf{PIQA} & \scriptsize\textbf{WG} & \scriptsize\textbf{ARC-E} & \scriptsize\textbf{ARC-C} & \scriptsize\textbf{Hella.} & \scriptsize\textbf{LogiQA} & \scriptsize\textbf{BoolQ} & \scriptsize\textbf{Lam.} & \\
\midrule
\textbf{MoE 12A2 L12} & 0.1B & 0.16B & 0.16B & 52.0 & 62.9 & 49.4 & 41.3 & 19.9 & 30.1 & 19.7 & 57.3 & 36.4 & 41.0 \\
\addlinespace[1pt]
\rowcolor{myred!50} \multicolumn{14}{@{}l@{}}{\textit{Width scaling}} \\
\aname 18A2 & 0.1B & 0.16B & 0.19B & \textbf{52.6} & 62.6 & 50.2 & 41.1 & \textbf{21.6} & 30.1 & 21.4 & 57.6 & 36.2 & 41.5 \\
\aname 24A2 & 0.1B & 0.16B & 0.22B & 52.5 & \textbf{63.3} & 50.9 & \textbf{41.8} & \textbf{21.6} & 29.3 & 19.7 & 59.6 & 36.1 & 41.6 \\
\aname 36A2 & 0.1B & 0.16B & 0.28B & \textbf{52.6} & \textbf{63.3} & \textbf{51.2} & 41.4 & 20.4 & \textbf{31.0} & \textbf{23.2} & \textbf{60.1} & \textbf{37.1} & \textbf{42.3} \\
\addlinespace[1pt]
\rowcolor{myblue!50} \multicolumn{14}{@{}l@{}}{\textit{Depth scaling}} \\
\aname L18 & 0.12B & 0.16B & 0.2B & 52.8 & 62.4 & 52.2 & \textbf{40.0} & 22.2 & 30.4 & 24.6 & 60.9 & 37.4 & 42.5 \\
\aname L24 & 0.13B & 0.18B & 0.24B & 55.1 & 62.8 & 52.8 & 39.0 & 23.2 & 30.7 & 25.8 & 60.3 & 38.8 & 43.2 \\
\aname L36 & 0.16B & 0.19B & 0.32B & \textbf{56.0} & \textbf{63.7} & \textbf{53.7} & 39.9 & \textbf{24.1} & \textbf{31.5} & \textbf{26.7} & \textbf{61.2} & \textbf{39.6} & \textbf{44.0} \\
\midrule
\textbf{MoE 64A8 L16} & 0.3B & 0.7B & 0.7B & 56.2 & 65.5 & 51.5 & 43.7 & 22.1 & 32.9 & 23.0 & 57.6 & 41.8 & 43.8 \\
\addlinespace[1pt]
\rowcolor{myred!50}  \multicolumn{14}{@{}l@{}}{\textit{Width scaling}} \\
 \aname 96A8 & 0.3B & 0.7B & 0.9B & 56.7 & \textbf{65.5} & 51.6 & 43.2 & 22.4 & \textbf{33.1} & 23.8 & 57.7 & 42.6 & 44.1 \\
\aname 128A8 & 0.3B & 0.7B & 1.16B & 57.2 & 65.1 & 51.7 & \textbf{44.7} & 22.7 & 32.4 & 24.5 & 57.8 & \textbf{43.4} & 44.4 \\
\aname 192A8 & 0.3B & 0.7B & 1.61B & \textbf{57.8} & 64.7 & \textbf{52.8} & 44.4 & \textbf{23.1} & 32.8 & \textbf{25.3} & \textbf{57.9} & 43.3 & \textbf{44.7} \\
\addlinespace[1pt]
\rowcolor{myblue!50}  \multicolumn{14}{@{}l@{}}{\textit{Depth scaling}} \\
\aname L24 & 0.38B & 0.75B & 0.98B & 57.2 & 65.3 & 52.9 & 42.7 & 23.8 & 33.2 & 26.4 & 59.3 & 42.8 & 44.8 \\
\aname L32 & 0.46B & 0.8B & 1.25B & 57.2 & \textbf{65.6} & 53.6 & 43.2 & 24.8 & 33.5 & 27.8 & 60.2 & 43.5 & 45.5 \\
\aname L48 & 0.61B & 0.9B & 1.8B & \textbf{58.3} & 65.2 & \textbf{54.3} & \textbf{44.0} & \textbf{25.8} & \textbf{33.9} & \textbf{28.8} & \textbf{61.0} & \textbf{44.2} & \textbf{46.2} \\
\bottomrule
\end{tabular}
\label{table:main-results1}
\vspace{-2mm}
\end{table*}

\begin{figure*}[t]
    \centering
    \includegraphics[width=1.\textwidth]{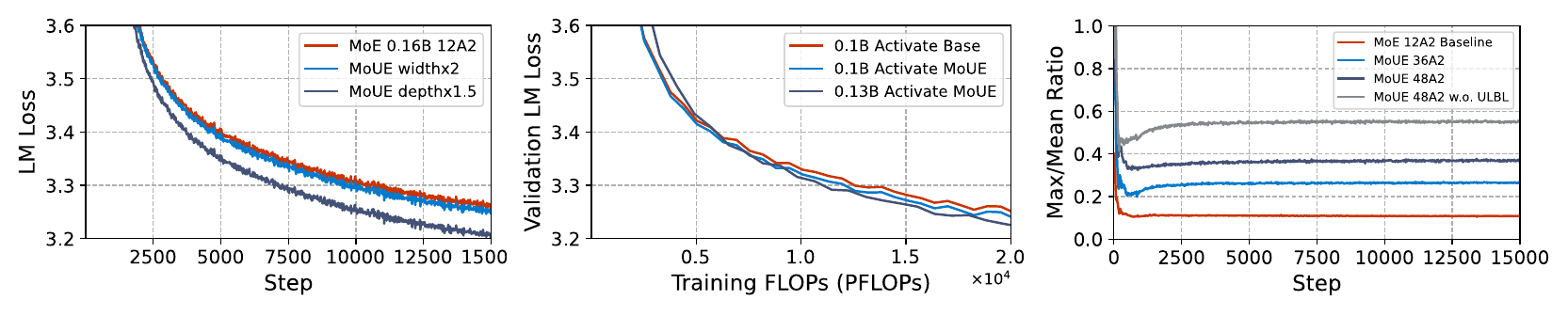}
    
    \caption{\textbf{Left:} LM training loss versus training steps. \textbf{Middle:} Validation loss versus training FLOPs. \textbf{Right:} Max/Mean ratio versus training steps (routing skew / load-balance indicator).}
    \label{fig:loss}
    
\end{figure*}
\subsection{Main Result}

\textbf{Width Expansion}
We investigate width expansion, where MoUE increases effective capacity by enlarging the Universal Expert pool without raising activated or total parameters; only virtual parameters grow, reflecting more compositional paths. As shown in Table~\ref{table:main-results1}, this yields consistent accuracy gains at fixed compute budgets: for MoE 12A2, Avg.\ improves from 41.0 to 42.3, and for MoE 64A8, from 43.8 to 44.7, with notable improvements on LogiQA, BoolQ, and LAMBADA. Performance scales further as VP increases, demonstrating that MoUE effectively translates additional virtual capacity into measurable improvements.

\textbf{Depth Expansion
}We next examine depth expansion, where MoUE scales the number of computation steps by reusing experts across layers, thus increasing depth with minimal growth in FFN parameters. This approach is highly parameter-efficient: for the 12A2 backbone, \aname L36 achieves Avg.\ 44.0 (+3.0) with TP only slightly above baseline; for 64A8, \aname L48 reaches Avg.\ 46.2 (+2.4) at TP=0.9B. Gains increase with model depth and scale, and notably, \aname L36 surpasses MoE 64A8 L16 in accuracy while using about half as activated and total parameters, demonstrating strong scalability under constrained budgets.


\textbf{Training Dynamics and Stability.}
Figure~\ref{fig:loss} summarizes the training-time behavior. Under width expansion, MoUE achieves better validation loss at the same FLOPs, indicating improved compute efficiency without increasing Act or TP. Under depth expansion, MoUE overtakes the baseline early and the gap continues to widen as training proceeds, suggesting a higher performance ceiling with longer training. Across settings, the Max/Mean ratio remains well-controlled, consistent with stable routing dynamics induced by staggered connectivity and UELB.
\begin{table*}[!ht]
\caption{Comprehensive evaluation of open-source MoE baselines and corresponding \aname settings. Each baseline is an open-source MoE model; below each are three \aname variants with their specific metrics.}
\vskip 0.15in
\setlength\tabcolsep{5pt}
\footnotesize
\centering
\small
\resizebox{0.98\textwidth}{!}{
\begin{tabular*}{0.98\textwidth}{@{\extracolsep{\fill}}@{}l cc cccccc cc cc c c@{}}
\toprule
\textbf{Model} & \textbf{Act.} & \textbf{VP.} & \textbf{ARC-c} & \textbf{ARC-e} & \textbf{GSM8K} & \textbf{HellaS.} & \textbf{HumanE.} & \textbf{NQ} & \textbf{Tri.QA} & \textbf{Wino.} & \textbf{MMLU} & \textbf{AVG} \\
\midrule
\rowcolor{myblue!50}  \multicolumn{13}{l}{\textit{JetMoE-8E, adapted by supervised fine-tuning.}} \\
Vanilla MoE & 2.2B & 8.0B & 64.4 & 79.0 & 44.7 & 78.7 & 42.7 & 20.4 & 49.5 & 56.6 & 47.7 & 53.7 \\
\aname\ 10A2 & 2.2B & 9.7B & \textbf{66.8} & 78.8 & 42.2 & 78.4 & 48.2 & 20.9 & 50.7 & 54.1 & 46.3 & 54.1 \\
\aname\ 12A2 & 2.2B & 11.3B & 65.1 & 79.4 & \textbf{45.1} & \textbf{80.3} & 42.7 & \textbf{21.4} & 50.5 & 56.8 & \textbf{48.2} & 54.4 \\
\aname\ 16A2 & 2.2B & 14.6B & 65.8 & \textbf{79.9} & 45.0 & 79.5 & \textbf{45.7} & 20.5 & \textbf{51.2} & \textbf{57.9} & 47.9 & \textbf{54.8} \\
\midrule
\rowcolor{myblue!50}  \multicolumn{13}{l}{\textit{OLMoE-64E, adapted by supervised fine-tuning.}} \\
Vanilla MoE & 1.3B & 6.9B & 57.0 & 76.7 & \textbf{36.3} & \textbf{62.1} & 24.2 & 18.1 & 45.3 & 52.3 & 47.1 & 46.5 \\
\aname\ 68A8 & 1.3B & 7.3B & 61.0 & 79.2 & 34.0 & 59.4 & 27.4 & 18.1 & 44.1 & 54.4 & 47.3 & 47.2 \\
\aname\ 72A8 & 1.3B & 7.7B & 61.4 & 79.5 & 35.5 & 58.7 & \textbf{29.3} & 18.0 & 44.5 & \textbf{56.5} & 46.0 & 47.7 \\
\aname\ 76A8 & 1.3B & 8.1B & \textbf{62.7} & \textbf{79.9} & 34.0 & \textbf{62.1} & 27.4 & \textbf{19.1} & \textbf{47.6} & 55.3 & \textbf{47.7} & \textbf{48.4} \\
\midrule
\rowcolor{myred!50}  \multicolumn{13}{l}{\textit{OLMoE-64E, adapted by continued pre-training.}} \\
Vanilla MoE & 1.3B & 6.9B & 37.3 & 40.2 & 31.0 & 26.3 & 13.4 & 9.6  & 18.4 & 49.1 & 46.3 & 30.2 \\
\aname\ 68A8 & 1.3B & 7.3B & 43.7 & 45.9 & \textbf{33.5 }& 26.2 & 14.0 & 9.3  & 22.0 & 49.9 & 47.5 & 32.4 \\
\aname\ 72A8 & 1.3B & 7.7B & 42.0 & 48.9 & 30.0 & \textbf{29.4} & 13.4 & \textbf{10.7} & \textbf{30.2} & 48.5 & \textbf{50.4} & 33.7 \\
\aname\ 76A8 & 1.3B & 8.1B & \textbf{50.9} & \textbf{58.2} & 31.5 & 26.4 & \textbf{14.6} & 8.0  & 22.9 & \textbf{50.4} & 47.1 & \textbf{34.4} \\
\bottomrule
\end{tabular*}
}
\label{tab:opensource-moe-main}
\end{table*}

\begin{figure*}[t]
    \begin{minipage}[c]{\textwidth}
    \centering
      \includegraphics[width=1\textwidth]{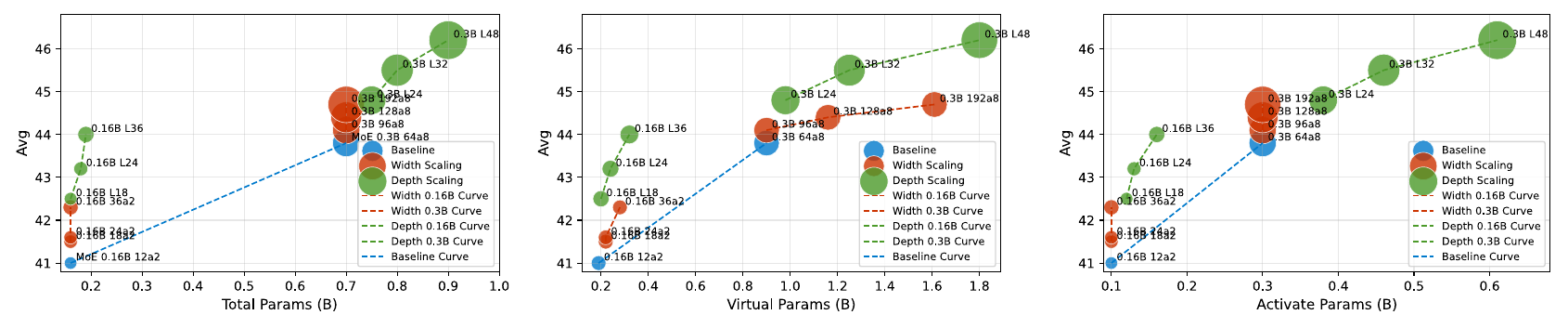}
      \caption{\textbf{Left:} Average accuracy versus total parameters, with point size representing activated parameters. \textbf{Middle:} Average accuracy versus virtual parameters, with point size representing activated parameters. \textbf{Right:} Average accuracy versus activated parameters.}\label{fig:scaling}
    \end{minipage}
\end{figure*}

\textbf{Scaling Analysis across TP, VP, and Act.}
Figure~\ref{fig:scaling} summarizes scaling trends across three budgets: total parameters (TP), virtual parameters (VP), and activated parameters (Act). \aname consistently achieves higher average accuracy than MoE at matched TP, VP, or Act. For TP, width scaling improves accuracy without increasing physical storage, while depth scaling yields even larger gains with only modest TP growth, demonstrating parameter efficiency. For VP, increasing virtual parameters leads to clear accuracy improvements, with depth expansion showing a steeper accuracy gain than width expansion, indicating more effective use of virtual capacity. For Act, \aname outperforms standard MoE at the same activated budget, and the advantage grows as Act increases, showing that MoUE leverages combinatorial computation to better utilize activated parameters.

\textbf{Progressive Warm-Start}
Training \aname from scratch is ideal but costly, so we investigate converting existing MoE checkpoints into \aname with minimal extra training and unchanged compute budgets. As shown in Table~\ref{tab:opensource-moe-main}, \aname consistently surpasses the original MoE baselines, with average gains of 1.1\% for JetMoE-8E and 1.9\% for OLMoE-64E, and up to +4.2\% on OLMoE-64E-CPT. Importantly, increasing the size of the universal expert pool leads to monotonic accuracy improvements, highlighting the benefit of recursive virtual width and the effectiveness of our progressive warm-start strategy.

\subsection{Ablation Studies}

\textbf{Setup.}
We evaluate ablations under matched training tokens and a fixed activated-parameter budget on the \aname 36A2 setting, and report validation perplexity. This isolates whether each component is necessary to make Virtual Width \emph{usable} rather than merely \emph{expressible}.

\textbf{Components Ablation.} Ablation results in Table~\ref{tab:abla} highlight the necessity of each key component in MoUE. Removing the \textbf{staggered rotational topology} leads to the largest degradation (+2.54 PPL), showing that Virtual Width requires structured connectivity for effective optimization; omitting \textbf{universal experts} also hurts performance, confirming the importance of cross-layer reuse. Among topology variants, our staggered rotational design consistently outperforms alternatives. For load balancing, replacing the proposed \textbf{UELB} with standard methods substantially increases perplexity (+0.97), and dropping connectivity normalization further degrades results, due to biased expert exposure and reduced path diversity. Probabilistic warmup helps early training. Finally, ablations on the router show that removing the \textbf{contextual pathway} or the \textbf{fast-weight update} both impair performance, indicating that coherent multi-step routing decisions—rather than shallow, stateless choices—are critical for leveraging recursive reuse. Together, these findings validate the necessity of structured connectivity, connectivity-aware load balancing, and stateful routing for realizing the benefits of Virtual Width.

\textbf{Warm-Start Warmup Ablation.}
The warm-start conversion relies on a logit-suppression curriculum $\beta(t)$ (Section~\ref{app:warmstart_full}): at the beginning, universal experts are strongly suppressed so the converted model behaves like the source MoE; $\beta(t)$ is then annealed to $0$ to gradually introduce reuse. If we ablate this warmup and turn on universal routing immediately, the pre-trained representations are disrupted and routing becomes unstable, which degrades convergence and downstream quality. This confirms that warm-starting \aname is not merely a parameter copy, but requires a controlled transition to the recursive reuse topology.

\begin{table}[t]
\centering

\caption{Eval perplexity for ablations on the \aname framework under matched training tokens and a fixed activated-parameter budget. ``w/o'' denotes removing the corresponding component from the full \aname model.}
\vskip 0.15in
\small
\setlength{\tabcolsep}{10mm}
\begin{adjustbox}{width=0.5\textwidth,center}
\begin{tabular}{@{}lc@{}}
\toprule
\textbf{Method} & \textbf{Perplexity} $\downarrow$  \\
\midrule
\aname 36A2 (full model) & 25.40 \\
\midrule
\multicolumn{2}{@{}l}{\textit{Connectivity / Expert Topology}} \\
w/o Staggered Rotational Topology & 27.94 {\small(\textcolor{red!50!black}{+2.54})} \\
w/o Universal Experts          & 26.05 {\small(\textcolor{red!50!black}{+0.65})} \\
\midrule
\multicolumn{2}{@{}l}{\textit{Routing Mechanism}} \\
w/o Contextual Pathway       & 25.83 {\small(\textcolor{red!50!black}{+0.43})} \\
w/o Fast-Weight Update             & 25.59 {\small(\textcolor{red!50!black}{+0.19})} \\
\midrule
\multicolumn{2}{@{}l}{\textit{Load Balancing Strategy}} \\
w/o UELB (standard load balance)                & 26.37 {\small(\textcolor{red!50!black}{+0.97})} \\
w/o Connectivity Normalization              & 25.75 {\small(\textcolor{red!50!black}{+0.35})} \\
w/o Probabilistic Warmup      & 25.66 {\small(\textcolor{red!50!black}{+0.26})} \\
\midrule
MoE 12A2 (iso-activate / full parameter)          & 26.25 {\small(\textcolor{red!50!black}{+0.85})} \\
\bottomrule
\end{tabular}
\end{adjustbox}
\label{tab:abla}

\end{table}

\subsection{Analysis}\label{sec.router.ana}

\begin{figure}[t]
\centering
\includegraphics[width=7.5cm]{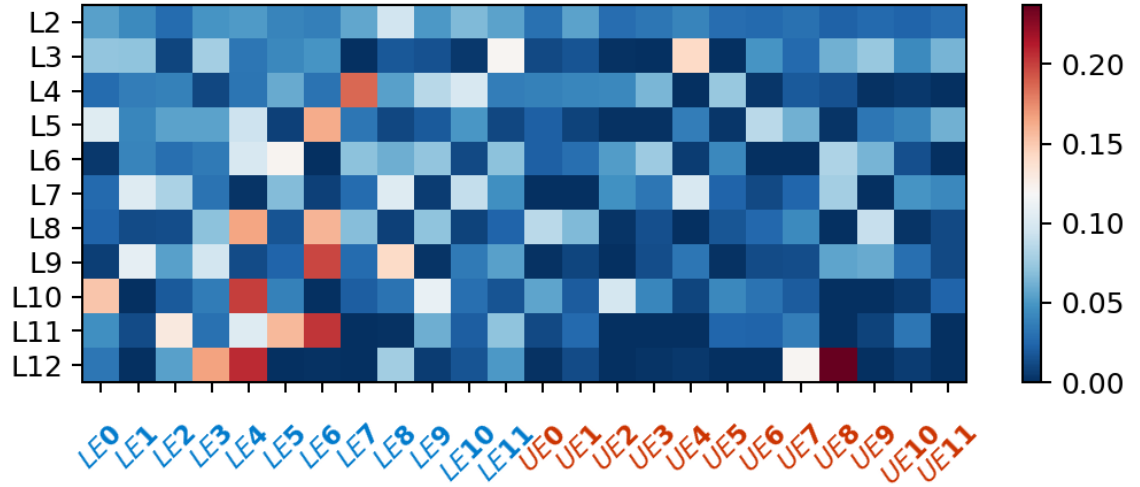}
\caption{Expert Utilization Heatmap. LE: Local Expert, UE: Universal Expert}
\label{fig:ue_heatmap}

\end{figure}

\textbf{Universal Expert Utilization.} We analyze the activation frequency of Universal Experts across layers to assess whether \aname achieves cross-layer reuse. As shown in Figure~\ref{fig:ue_heatmap}, UELB enables Universal Experts to be activated broadly in both shallow and deep layers, with usage patterns comparable to layer-private experts. In deeper layers, private experts show specialization but are still selected by multiple layers, demonstrating effective parameter sharing.

\textbf{UELB Improves Load Balance Dynamics.} To assess training stability, we monitor routing skew via the Max/Mean ratio. Figure~\ref{fig:loss} (right) shows that without UELB, routing quickly collapses, while UELB maintains low skew throughout. By normalizing for expert exposure and enforcing depth-wise balance, UELB ensures that Virtual Width translates to capacity rather than repeated paths.

\textbf{Effect of Universal Expert Pool Size.}
Figure~\ref{fig:expert_scaling} shows that increasing the number of Universal Experts (UEs) per layer in the warm-start setting improves performance up to an optimal point (12 UEs), after which further increases degrade results. This demonstrates that moderate sharing effectively exploits cross-layer redundancy, but excessive sharing can harm layer-specific specialization.

\textbf{Domain Specialization of Universal Experts.}
As shown in Figure~\ref{fig:expert_domain}, universal experts develop distinct domain preferences (e.g., some specialize in Math, others in Code), indicating spontaneous functional specialization and validating the Universal Router's ability to dynamically match tokens to relevant experts.

\textbf{Layer-wise Utilization of Universal Experts.}
In Figure~\ref{fig:expert_sft_ratio}, that UEs are used sparsely in early layers but much more in deeper layers, suggesting the model learns to rely on LEs for low-level features and on UEs for high-level reasoning.

\textbf{Topology Impact and Functional Circuits.}
Table~\ref{tab:topology_ablation} shows that the staggered rotational topology yields the best performance and forms stable expert circuits, while other topologies introduce instability or disordered expert flow, highlighting the importance of structured connectivity for compositional capacity.
\begin{table}[t]
\centering

\caption{Ablation study on connectivity topologies. We compare our \textbf{Staggered Rotational} design against alternative connectivity schemes under matched training tokens and a fixed activated-parameter budget.}
\vskip 0.15in
\small
\setlength{\tabcolsep}{3mm}
\begin{adjustbox}{width=0.5\textwidth,center}

\begin{tabular}{@{}lcc@{}}
\toprule
\textbf{Topology Scheme} & \textbf{Perplexity} $\downarrow$ & \textbf{Circuit Pattern} \\
\midrule
\rowcolor{myblue!10} \textbf{Staggered Rotational (Ours)} & \textbf{25.40} & \textbf{Re-entrant Loops} \\
\midrule
Forward Sliding Window & 25.87 {\small(\textcolor{red!70!black}{+0.47})} & Sequential Drift \\
Reverse-Order Connection & 25.93 {\small(\textcolor{red!70!black}{+0.53})} & Disordered Flow \\
Sandwich Architecture & 25.61 {\small(\textcolor{red!70!black}{+0.21})} & Rigid Stratification \\
Full All-to-All (Unconstrained) & 27.94 {\small(\textcolor{red!70!black}{+2.54})} & Random / Unstable \\
\bottomrule
\end{tabular}
\end{adjustbox}
\label{tab:topology_ablation}
\vspace{-2mm}
\end{table}

\section{Related Work}

\textbf{Sparse MoE models} ~\citep{DBLP:journals/corr/ShazeerMMDLHD17} employ learnable gating to activate a subset of experts per token, enabling efficient scaling~\citep{DBLP:journals/corr/abs-2006-16668, fedus2022switch, jiang2024mixtral, gu2025elasticmoeunlockinginferencetime,chen2025mixture}. 
Routing approaches span load balancing~\citep{fedus2022switch, DBLP:journals/corr/abs-2006-16668}, hash-based~\citep{DBLP:journals/corr/abs-2106-04426}, expert-choice~\citep{zhou2022mixture}, BASE layers~\citep{DBLP:journals/corr/abs-2103-16716}, and soft mixtures~\citep{puigcerver2024sparsesoftmixturesexperts}, but do not address the recursive, multi-layer exposure of universal experts in MoUE. 

\textbf{Universal Transformers}~\citep{dehghani2019universal,lan2019albert,ng2024loopneuralnetworksparameter} and earlier works on recurrence~\citep{braitenberg1986vehicles,gers2000lstm} introduce recurrence via shared transformations, with recent research revisiting recurrence for implicit reasoning~\citep{deng2023implicit,hao2024traininglargelanguagemodels} and test-time scaling~\citep{geiping2025scalingtesttimecomputelatent}. Further Discussion in Appendix~\ref{app.rw}.

\section{Conclusion}

In this work, we introduce Mixture of Universal Experts (\aname), a novel framework that scales Mixture-of-Experts models through cross-layer expert reuse and virtual width. By leveraging a structured connectivity topology, connectivity-normalized load balancing, and a stateful routing mechanism, \aname significantly expands model capacity without increasing the activated parameter or compute budget. Extensive experiments demonstrate that \aname consistently outperforms standard MoE baselines in both width and depth expansion settings, achieving greater efficiency and scalability.

\bibliography{icml_2025}
\bibliographystyle{icml2025}

\newpage
\appendix

\section{Related Work} \label{app.rw}

\subsection{Mixture-of-Experts and Conditional Computation}

Sparse MoE models activate a small subset of experts per token to scale total parameters without proportional compute growth. The sparsely-gated MoE layer \citep{DBLP:journals/corr/ShazeerMMDLHD17} introduced learnable gating for conditional computation. Subsequent work scale this paradigm to hundreds of billions of parameters \citep{DBLP:journals/corr/abs-2006-16668, fedus2022switch, jiang2024mixtral} and demonstrate strong open-source performance with a moderate number of experts. Recent work has explored finer-grained expert specialization \citep{dai2024deepseekmoeultimateexpertspecialization} and unified scaling laws for MoE models \citep{krajewski2024scalinglawsfinegrainedmixture}. Despite this progress, standard designs typically partition experts by layer, making total expert memory scale linearly with the number of experts and limiting the reuse of learned operators across depth.

\subsection{Routing and Load Balancing}
Training MoE models requires careful routing design to avoid expert collapse and improve utilization. Prior work commonly uses auxiliary load-balancing losses \citep{fedus2022switch, DBLP:journals/corr/abs-2006-16668}, capacity constraints, and routing regularization. Alternative routing strategies include hash-based assignment \citep{DBLP:journals/corr/abs-2106-04426}, expert-choice routing \citep{zhou2022mixture}, and BASE layers \citep{DBLP:journals/corr/abs-2103-16716}. Soft mixture approaches \citep{puigcerver2024sparsesoftmixturesexperts} relax discrete routing entirely. \textsc{MoUE} introduces a structurally new challenge: universal experts are exposed to multiple layers by design, which breaks the uniform prior assumption behind standard load balancing. Our UELB objective explicitly normalizes for connectivity and stabilizes routing under recursive reuse.

\subsection{Recurrence and Parameter Reuse Across Depth}
Universal Transformers~\citep{dehghani2019universal} explore recurrence across depth by reapplying a shared transformation, offering an inductive bias of iterative refinement but often requiring additional width to maintain capacity. \aname retains the iterative refinement perspective while relaxing strict parameter tying: it reuses a shared \emph{set} of experts with learned routing over a constrained topology, enabling richer compositions than repeatedly applying an identical block.

\section{General Framework}
\label{app:general_framework_full}

This appendix contains the full version of Section~\ref{sec:moue.general}, including an extended discussion condensed in the main paper.

\subsection{General Framework}

To transcend the architectural rigidities of standard MoEs, we propose the \aname. This framework reinterprets the deep neural network not as a static stack of heterogeneous layers, but as a recursive computational process over a unified, globally accessible expert space.

\textbf{Recursive Formulation via Connectivity Topology}
Let us consider a Transformer of conceptual depth $L$. We decompose the expert space into two disjoint sets: a set of layer-specific \textit{local experts} $\mathcal{E}^{\text{local}}_{\ell}$ and a globally shared pool of \textit{Universal Experts} (UEs), $\mathcal{E}^{\text{u}}$. The total expert universe is denoted by $\mathcal{E}^{\star} = (\bigcup_{\ell} \mathcal{E}^{\text{local}}_{\ell}) \cup \mathcal{E}^{\text{u}}$.

The architecture of \aname is defined by a flexible \textbf{Connectivity Topology Mapping} $\mathcal{C}: \mathcal{E}^{\star} \times \{1, \dots, L\} \rightarrow \{0, 1\}$, which prescribes the adjacency matrix of the bipartite graph between computational layers and the expert pool. Concretely, for a local expert $E \in \mathcal{E}^{\text{local}}_{\ell}$ we set $\mathcal{C}(E,\ell)=1$ and $\mathcal{C}(E,m)=0$ for $m\ne \ell$; for a universal expert $u\in \mathcal{E}^{\text{u}}$, the topology decides the set of layers where $\mathcal{C}(u,\ell)=1$. The forward propagation at layer $\ell$ is modeled as a conditional query over this topology:
\begin{equation}
\begin{split}
    h^{(\ell)} &= h^{(\ell-1)} + \\
    &\sum_{E_i \in \mathcal{E}^{\star}} \mathcal{C}(E_i, \ell) \cdot \pi^{(\ell)}_i(h^{(\ell-1)}) \cdot E_i(h^{(\ell-1)}).
\end{split}
\end{equation}
Crucially, for universal experts, the mapping $\mathcal{C}(u, \ell)$ can be active across multiple layers. This enables the router to learn a dynamic policy that reuses parameters at varying levels of semantic abstraction, effectively decoupling the parameter space from the computational depth.

\textbf{Theoretical Capacity: The Virtual Width Hypothesis}
The defining advantage of the \aname framework is the exponential expansion of model capacity through combinatorics, a property we term \textbf{Virtual Width}. In a standard MoE, the number of functional paths is constrained by the physical width of the layer. In contrast, under a fully connected \aname topology where universal experts are accessible at every step, the router constructs a computational path by selecting any $k$ experts from $N_u$ at each of the $L$ steps.

The magnitude of the functional path space $\mathcal{T}$ scales according to:
\begin{equation}
    |\mathcal{T}|_{\text{\aname}} \approx \left( \binom{N_u}{k} \right)^L.
\end{equation}
This formulation highlights that the functional path space can grow exponentially with depth, even though parameter storage grows only linearly with the number of universal experts. In other words, \aname can support a much larger set of composable expert paths without proportionally increasing physical expert memory.

We describe this as a \textbf{free lunch} in \emph{physical expert memory}: once the activated-parameter budget is fixed, increasing reuse opportunities can increase effective capacity without requiring expert memory to grow in lockstep. This does not mean optimization is free---the rest of this section focuses on the constraints needed to make this capacity usable in practice.

\textbf{Depth--Width Exchange via Loop MoE}
Virtual Width can be interpreted as a \textbf{depth--width exchange}. Consider a window of $W$ consecutive layers that share the same accessible universal expert set (i.e., the same expert mask). If we conceptually \emph{tie} this window into a loop, then routing within the window resembles a \emph{Loop MoE}: instead of allocating distinct experts for each layer, the model repeatedly queries a reusable expert pool over multiple steps. Increasing the number of steps (depth) enlarges the space of expert compositions, playing a role analogous to increasing width in a conventional MoE. This exchange is the core of the fourth-dimension scaling: we turn additional depth into additional effective width by reuse, while keeping the activated expert budget per token unchanged.

\textbf{The Optimization Bottleneck}
However, this theoretical capacity comes at the cost of severe \textbf{Optimization Complexity}. If the connectivity $\mathcal{C}$ permits all-to-all access, the router confronts a massive search space at every layer, creating a dimensionality problem of size $L \times N_u$.

This highlights an important modeling perspective: \aname induces a rapidly growing space of possible expert compositions over depth, which can translate into increased effective capacity without requiring the physical expert memory to grow at the same rate. This is an \emph{idealized} statement about combinatorics; in practice, the challenge is to make the router reliably discover diverse, useful paths.

This introduces two distinct failure modes. First, as $N_u$ grows to increase capacity, the router must discriminate among a vast number of experts at every step. This dilutes the gradient signal, leading to high variance and training instability. Second, without structural constraints, recursive reuse can precipitate a ``Deep Equilibrium Trap,'' where the model converges to a fixed-point solution by selecting the same subset of generic experts repeatedly. This repetitive dependency prevents the model from exploiting the combinatorial potential of Virtual Width, effectively collapsing the deep network into a shallow recurrent loop. Thus, realizing the benefits of combinatorial path capacity requires structural constraints that render optimization tractable.

\section{Staggered Connectivity}
\label{app:staggered_full}
\begin{algorithm}[t]
\caption{Progressive Warm-Start (MoE $\rightarrow$ \aname)}
\label{alg:warmstart}
\begin{algorithmic}[1]
\STATE \textbf{Input:} MoE checkpoint; topology $(G,W,s)$; warmup schedule $\beta(t)$
\STATE Initialize $\mathcal{E}^{\text{u}}$ by copying selected MoE experts; keep others as local experts
\FOR{$t=1$ to $T$}
    \STATE Compute local routing logits $z^{(\ell)}_{\text{local}}$
    \STATE Compute universal routing logits $z^{(\ell)}_{\text{u}} - \beta(t)$
    \STATE Route with combined logits; apply staggered connectivity $\mathcal{C}$
    \STATE Optimize task loss + UELB (connectivity-normalized auxiliary loss)
\ENDFOR
\STATE \textbf{Output:} Warm-started \aname checkpoint
\end{algorithmic}
\end{algorithm}
This appendix contains the full version of Section~\ref{sec:moue.staggered}, including details that are condensed in the main paper.

\subsection{Structural Optimization: Staggered Connectivity}

To bridge the gap between the theoretical potential of \aname and its practical optimization difficulties, we propose the \textbf{Staggered Rotational Topology}. This design acts as a structural regularizer on the connectivity map $\mathcal{C}$, enforcing a controlled evolution of the expert space.

\textbf{Topology Design Space.}
The connectivity map $\mathcal{C}$ admits many plausible designs (e.g., all-to-all, fixed window sharing, random sparsification, block-circulant masks). In our exploration, we found that a compact, ring-based construction strikes the best balance between expressivity and trainability. The key is to make reuse \emph{structured}: the router should not face an unstructured global search at every layer, yet the reachable set should evolve with depth to avoid trivial loops.

\textbf{Two-Level Expert Ring (Coarse Shared Window + Fine Private Window).}
We structure the expert universe on a two-level ring. First, a \emph{large} sliding window defines the \emph{set of experts visible to a connectivity group} of $G$ consecutive layers: all layers in the group share the same reachable set. Second, within this group-visible set, a \emph{small} sliding window assigns a disjoint subset as \emph{layer-private} experts for each layer; the remaining experts form the shared UE bucket that can be reused across the group (and, via rotation, across depth). This construction gives a simple specialization--reuse decomposition while keeping masks easy to implement and analyze.

\textbf{Staggered Rotational Topology}
We structure the universal expert pool $\mathcal{E}^{\text{u}}$ as a logical ring buffer (experts are indexed on a ring). We first group consecutive layers into \textbf{connectivity groups} of size $G$ (e.g., $G=3$ layers share the same universal expert mask). Let $g(\ell)=\lfloor(\ell-1)/G\rfloor$ be the group index of layer $\ell$. We enforce a sliding window over the expert ring with expert-window size $W$ and stride $s$ (shift per group). The connectivity function for universal experts is strictly constrained as:
\begin{equation}
\begin{split}
    \mathcal{C}&(E_j, \ell) = \\ 
    &\mathbb{I}\left(j \in \left[ g(\ell) \cdot s, \dots, g(\ell) \cdot s + W - 1 \right] \pmod{N_u} \right).
\end{split}
\end{equation}
This imposition creates a band-diagonal structure on the expert-layer adjacency matrix, ensuring that the set of accessible experts shifts progressively with network depth, while remaining constant within each $G$-layer group.

\textbf{Implementation: Layer-wise Reachability Lists}
In implementation, we instantiate $\mathcal{C}$ as a per-layer \emph{reachability allow-list} over global expert IDs. Each MoE layer masks out experts that are not in its allow-list by setting their routing logits to $-\infty$ before Top-$k$ selection, so the allow-list only controls reachability and does not encode any ordering or preference. A connectivity group corresponds to a contiguous block of $G$ layers that share an identical allow-list; the staggered rotation is implemented by shifting the universal window start by stride $s$ between adjacent groups. To align with our notation, we partition global IDs into a contiguous local block per layer plus a universal block shared across layers, so that $N_u$, $W$, $G$, $s$, and Top-$k$ map directly to a concrete mask construction.

\textbf{From Window Sharing to Staggered Sharing}
The simplest window-sharing scheme fixes a single universal mask and applies it to all layers. This directly implements the depth--width exchange, but scales poorly when extended globally: the router repeatedly faces a large, unstructured search space, and reuse can collapse to trivial loops. The \textbf{staggered} design keeps $G$ (layers per group), $W$ (reachable universal experts), and the per-token top-$k$ fixed, but \emph{rotates} the reachable experts across groups via stride $s$. This retains the local loop property (reuse within a small depth range), while reducing global complexity and improving optimization stability.

\textbf{Solving the Optimization Bottleneck}
This topological constraint directly addresses the optimization bottleneck discussed above by transforming the global routing problem into a sequence of localized sub-problems. By reducing the active search space at any layer from $N_u$ to $W$ (where $W \ll N_u$), we ensure that the router learns to discriminate only among a locally relevant subset of experts.

Furthermore, the sliding window mechanism serves to break repetitive sequential dependencies. Since the window shifts, a specific expert $E_j$ appears in different relative positions within the window at different layers. This forces the expert to function as a ``position-agnostic operator,'' robust to varying contexts. Despite this reduction in instantaneous search space, the Virtual Width remains exponential. The number of unique paths is preserved as $|\mathcal{T}| \approx \prod_{\ell} \binom{W}{k}$, allowing the model to retain massive capacity scaling while ensuring optimization feasibility.

\section{Universal Expert Load Balance (UELB)}
\label{app:uelb_full}

This appendix contains the full version of Section~\ref{sec:moue.uelb}. Additional implementation details on group-wise estimation and scale correction are provided in Appendix~\ref{app:exp_setup_full}.

\subsection{Universal Expert Load Balance (UELB)}

%

While the Staggered Topology resolves structural complexity, it introduces a severe statistical imbalance which we identify as the \textbf{Heterogeneous Exposure Problem}. This represents a core theoretical conflict between recursive architectures and standard optimization objectives.

\textbf{The Principle of Architectural Bias Correction}
The standard Load Balancing Loss (Eq. 2) operates on the assumption of Homogeneous Prior Utility: it presumes that in an initialized state, every expert should be selected with uniform probability $1/N$. The \aname topology, however, explicitly violates this assumption. A universal expert $E_u$ is structurally connected to $c_u$ layers, while a local expert is connected to only one.

If the standard LBL is applied directly, the loss function aggregates probability mass for $E_u$ from $c_u$ distinct sources. It misinterprets this architecturally induced accumulation as ``over-utilization'' and heavily penalizes the universal experts. This results in a gradient bias that is not a penalty on suboptimal routing, but a penalty on the architecture itself. The optimization process consequently suppresses universal experts to equalize their total usage with local experts, effectively dismantling the shared pool and negating the Virtual Width advantage.

\textbf{Universal Expert Load Balance Strategy}
To rectify this structural misalignment, we propose the UELB strategy, which introduces a \textbf{Connectivity-Normalization Principle}. We posit that load balancing must be enforced relative to topological opportunity rather than absolute usage.

We formulate the UELB objective as a composite function:
\begin{equation}
\begin{split}
    \mathcal{L}_{\text{UELB}} =  
    &\alpha_{\text{loc}} \sum_{\ell} \sum_{i \in \mathcal{E}^{\text{loc}}} \bar{f}_i \bar{P}_i \\ &+ \alpha_{\text{u}} \sum_{j \in \mathcal{E}^{\text{u}}} \frac{1}{c_j} \left( \sum_{\ell=1}^L \mathcal{C}(E_j, \ell) \cdot \bar{f}_j^{(\ell)} \bar{P}_j^{(\ell)} \right),
\end{split}
\end{equation}
where $c_j = \sum_\ell \mathcal{C}(E_j, \ell)$ represents the topological degree of expert $j$, $\bar{f}_i$ and $\bar{P}_i$ denote the average routing frequency and expected load (or utilization probability) for expert $i$ at the local branch, respectively. The term $1/c_j$ acts as a theoretical correction factor that transforms the cumulative probability sum back into an \textit{average layer-wise utilization rate}. By optimizing this objective, we ensure that the router is penalized only if an expert is over-utilized \textit{within the contexts where it is available}. This effectively decouples ``architectural popularity'' (designed by topology) from ``routing popularity'' (learned from data), enabling the model to fully leverage universal expert recurrence without gradient interference.

\textbf{How We Estimate $\bar{f}$ and $\bar{P}$ in Practice}
In implementation, we estimate $\bar{f}$ and $\bar{P}$ at the granularity of a \textbf{connectivity group}: a set of layers that share the same universal expert mask (i.e., identical $\mathcal{C}(\cdot,\ell)$). Concretely, for each group we collect, across all tokens in the batch and all layers in the group, (i) the router probabilities over the reachable experts and (ii) the discrete token assignments (Top-$k$ indices) used to compute per-expert token counts. We then concatenate these tensors and compute a single Switch-style auxiliary loss for the group.

\textbf{Probabilistic Warmup as Regularization}
Complementing this objective, we address the exploration dilemma inherent to the expanded Virtual Width. Early in training, the router tends to collapse to local minima before exploring the full combinatorial potential. We introduce a probabilistic warmup bias added to the universal logits, which boosts exploration of the universal pool in the initial phase and is annealed away later. We implement warmup by adding a shared-expert logit bias to the universal bucket before the softmax: at training progress $t$, we multiply the pre-softmax mass of universal experts by a scalar ratio $\rho(t)$ and equivalently add $b(t)=\log \rho(t)$ to their logits.

\textbf{Calibrating $\alpha_{\text{loc}}$ and $\alpha_{\text{u}}$}
Because group-wise aggregation changes the scale of the auxiliary loss, we calibrate $\alpha_{\text{loc}}$ and $\alpha_{\text{u}}$ as deterministic functions of (i) the number of layers per connectivity group, (ii) the number of shared experts, and (iii) Top-$k$, so the auxiliary-loss magnitude transfers across model sizes and reuse topologies.

\textbf{Strict Control Baseline}
To isolate the effect of UELB from other factors, our primary control baseline uses the same topology $\mathcal{C}$ and the same router, and only swaps the auxiliary objective: UELB vs.\ the standard load-balancing loss.

\section{Universal Router}
\label{app:router_full}

This appendix contains the full version of Section~\ref{sec:moue.router}, including connectivity-group state sharing and the stateless-router control.

\subsection{Universal Router}

While the Staggered Topology effectively constrains the instantaneous search space to a tractable subset, navigating the resulting high-dimensional combinatorial manifold requires a routing mechanism capable of context awareness beyond immediate local representations. Standard stateless routers, which treat layer-wise decisions as independent events, inherently fail to capture the recursive dependencies and long-range correlations distinct to the \aname framework. To bridge this gap, we propose the Universal Router, a dynamic decision-making module designed to maintain trajectory coherence across the depth of the network by integrating historical state information into the expert selection.

\textbf{Dual-Pathway Routing Mechanism}
To navigate the complex path space defined in Section 2, we introduce a stateful Universal Router. We decompose the routing process into two distinct pathways:
\begin{equation}
    z^{(\ell)} = \underbrace{h^{(\ell)} W_g + b_g}_{\text{Semantic Pathway}} + \underbrace{\beta \cdot \text{Sim}(h^{(\ell)}, U^{(\ell)})}_{\text{Contextual Pathway}}.
\end{equation}
The \textbf{Semantic Pathway} handles the static, affine matching between token features and expert capabilities. The \textbf{Contextual Pathway} incorporates a dynamic state matrix $U^{(\ell)}$ (depth fast-weights) to capture trajectory history. We compute similarity using a projected key $k^{(\ell)} = h^{(\ell)} W_k \in \mathbb{R}^{d_k}$ and $\text{Sim}(h^{(\ell)}, U^{(\ell)})_j = \langle u_j^{(\ell)}, k^{(\ell)} \rangle$. Here, $K$ is the matrix of projected keys for all tokens in the batch and $p^*$ denotes a detached routing target distribution.

\textbf{Online State Update}
We implement fast weights as a forward-only state and update them online after routing. Concretely, we detach $U^{(\ell)}$ from autograd and perform a no-gradient update using the current token keys $K$ and a detached routing target $p^*$ (either masked router probabilities or the hard routing map). Let $\hat{p}=\operatorname{softmax}(K U^\top)$ be the contextual prediction and $\delta=\hat{p}-p^*$. We update with:
\begin{equation}
    U^{(\ell)} \leftarrow U^{(\ell)} - \eta_\ell \cdot \frac{\delta^\top K}{N_{\text{tok}}},
\end{equation}
where $N_{\text{tok}}$ is the number of tokens used for normalization, and $\eta_\ell$ can optionally decay with depth. Importantly, we do not backpropagate gradients through this update, keeping activation memory unchanged relative to the MoE baseline.

\textbf{State Sharing Within Connectivity Groups}
To align the router state with the reuse topology, we share the fast-weight state across layers that have the same reachability mask. Layers with identical allow-lists reuse a single $U$ instance, so the contextual pathway tracks trajectories within a fixed local subgraph of reachable experts rather than resetting per layer.

\textbf{Stateless Router Baseline}
Our stateless-router control disables the contextual pathway by setting $\beta=0$ and skipping the fast-weight update, while keeping the same topology, expert parameters, and training setup. This baseline isolates whether trajectory state is necessary for navigating the reuse graph.

\section{Progressive Warm-Start}
\label{app:warmstart_full}

This appendix contains the full version of the progressive warm-start procedure summarized in the main paper.

\subsection{Progressive Warm-Start}

While training \aname from scratch yields optimal topology-aware representations, the computational cost can be prohibitive. To leverage existing high-performance checkpoints, we propose a \textbf{Progressive Transformation Strategy} that converts a standard pre-trained MoE into \aname via a three-stage warm-start process. This approach relies on the observation that standard MoEs naturally develop redundant functional clusters across layers (as shown in Figure~\ref{fig:expert_similarity}).

\textbf{Spectral Initialization of the Universal Pool}
To leverage the capabilities of pre-trained standard MoE models and accelerate convergence, we propose a streamlined \textbf{Progressive Transformation Strategy}. This approach initializes the \aname architecture by inheriting weights from a source MoE and employing a curriculum-based transition to the recursive topology.

\textbf{1. Heuristic Initialization of Universal Experts}
We construct the initial Universal Expert pool $\mathcal{E}^{\text{u}}$ by selecting a representative subset of experts from the pre-trained model. Rather than complex clustering, we employ a simplified heuristic: experts with the highest activation rates across intermediate layers (indicating high generalizability) are cloned to initialize $\mathcal{E}^{\text{u}}$. The remaining experts serve as the initial Local Experts $\mathcal{E}^{\text{local}}$, preserving the layer-specific semantic knowledge of the original model.

\textbf{2. Curriculum Routing Warmup via Logit Suppression}
To integrate the Universal Experts without disrupting the pre-trained feature space, we introduce a dedicated router projection $W_{r}^{\text{u}}$ at each layer, running in parallel with the preserved local router $W_{r}^{\text{local}}$. To manage the transition, we employ a \textbf{Logit Suppression Strategy} that imposes a time-dependent negative bias on the universal branch.

Formally, let $z^{(\ell)} \in \mathbb{R}^{|\mathcal{E}^{\star}|}$ denote the routing logits at layer $\ell$. Here, $z^{(\ell)}_i$ for expert $E_i$ is computed as:
\begin{equation}
    z^{(\ell)}_i =
    \begin{cases}
        h^{(\ell-1)} \cdot W_{r}^{\text{local}}[\cdot, i] & \text{if } E_i \in \mathcal{E}^{\text{local}}_{\ell} \\
        h^{(\ell-1)} \cdot W_{r}^{\text{u}}[\cdot, i] - \beta(t) & \text{if } E_i \in \mathcal{E}^{\text{u}}
    \end{cases}
\end{equation}
where $\beta(t) \geq 0$ is a scalar bias controlled by a curriculum scheduler. The final gating probabilities are obtained via $\pi^{(\ell)} = \text{Softmax}(z^{(\ell)})$.

At the beginning of training ($t=0$), we initialize $\beta(0) \to \infty$ (e.g., $10^4$), which suppresses the activation probability of $\mathcal{E}^{\text{u}}$ to near-zero, forcing the model to rely solely on $\mathcal{E}^{\text{local}}_{\ell}$ and preserving the original behavior. As training proceeds, $\beta(t)$ is linearly annealed to $0$, allowing the gradients to gradually warm up the parameters of the universal router $W_{r}^{\text{u}}$ and smoothly integrate the recursive topology.

\section{Experimental Setup}
\label{app:exp_setup_full}
\begin{figure}[t]
\centering
\includegraphics[width=6cm]{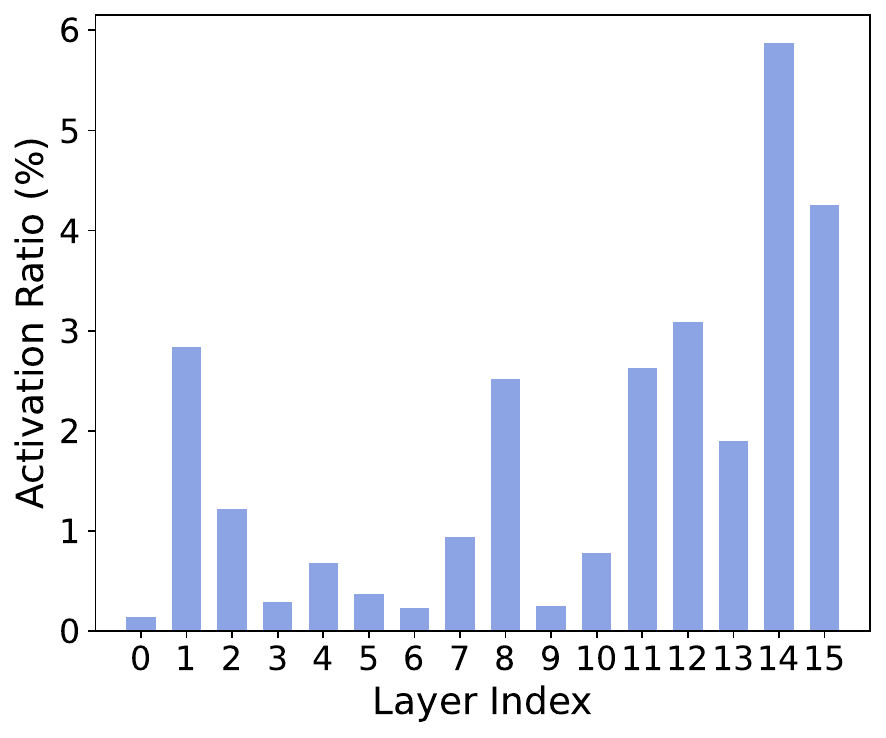}
\caption{Token Routing Proportion to Universal Experts per Layer after SFT warm-start training.}
\label{fig:expert_sft_ratio}

\end{figure}

\begin{figure}[t]
\centering
\includegraphics[width=6cm]{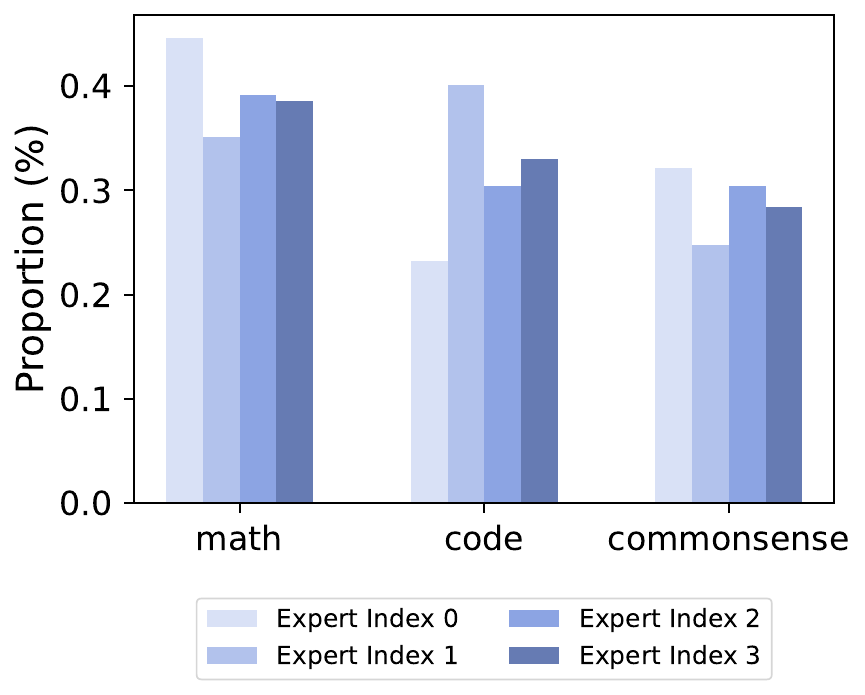}
\caption{Activation differences of universal experts across domains. Experiments are conducted with 4 universal experts per layer.}
\label{fig:expert_domain}

\end{figure}
\begin{figure}[t]
\centering
\includegraphics[width=6cm]{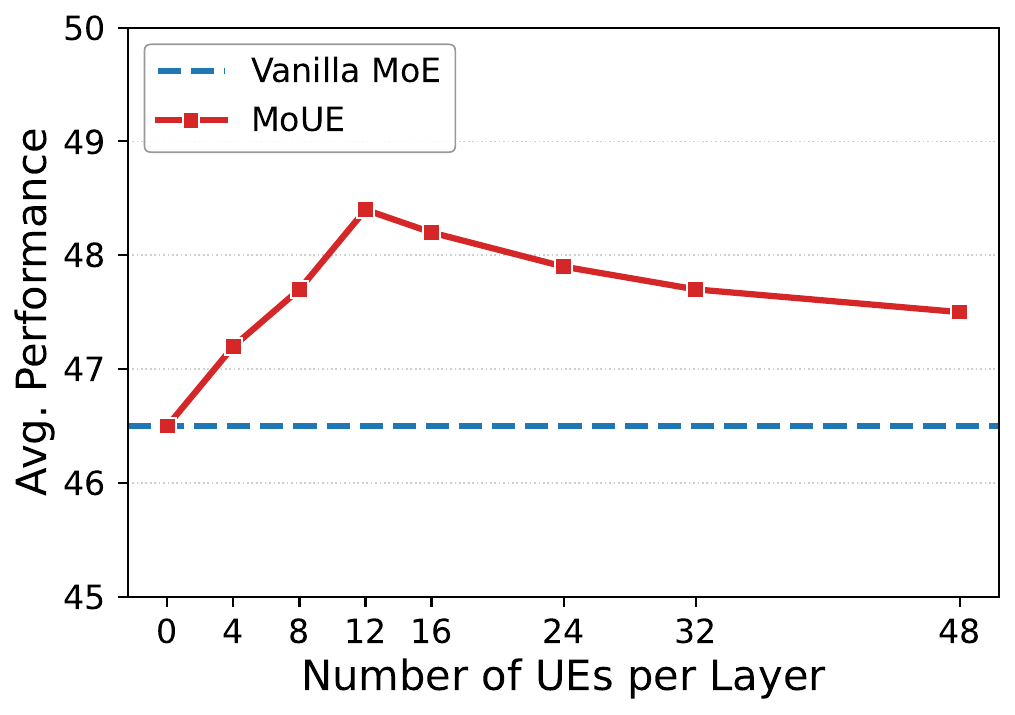}
\caption{Effect of the number of universal experts in warm-start experiments.}
\label{fig:expert_scaling}

\end{figure}

This appendix contains the full version of Section~\ref{sec:exp.setup}.
\subsection{Experimental Setup}

\textbf{Scope of this appendix.}
Data/training protocol and iso-parameter / iso-compute definitions are already given in Section~\ref{sec:exp.setup}; we do not repeat them here.

\textbf{Evaluation.} We employed the lm-evaluation-harness~\cite{eval-harness} to evaluate our models. For common sense and reading comprehension tasks, we report 0-shot accuracy results for SciQ~\cite{sciqa}, PIQA~\cite{piqa}, WinoGrande (WG)~\cite{WinoGrande:conf/aaai/SakaguchiBBC20}, ARC Easy (ARC-E)~\cite{clark2018think}, and 10-shot HellaSwag (Hella.)~\cite{HellaSwag:conf/acl/ZellersHBFC19}, alongside 25-shot accuracy for ARC Challenge (ARC-C)~\cite{arcChallenge:journals/corr/abs-1803-05457}. In the assessments of continued QA and text understanding, we report 0-shot accuracy for LogiQA~\cite{liu2020logiqa}, 32-shot BoolQ~\cite{clark2019boolq}, and 0-shot LAMBADA (Lam.)~\cite{paperno2016lambada}. All reported results are calculated with the mean and stderr of multiple experiments.

\textbf{Baseline.} Following the architecture of Qwen-3, we evaluate two MoE backbones, both using standard top-$k$ routing with experts local to each layer:
\begin{itemize}[leftmargin=*]
    \item \textbf{MoE-160M:} 18 layers, hidden size 768, 12 attention heads, 4 query groups, 12 experts per layer, top-$k=2$, and MoE FFN hidden size 192.
    \item \textbf{MoE-700M:} 16 layers, hidden size 1536, 16 attention heads, 4 query groups, 64 experts per layer, top-$k=8$, and MoE FFN hidden size 96.
\end{itemize}
All non-\aname components (tokenization, optimizer, learning-rate schedule, data sampling, and evaluation harness) are kept identical between MoE and \aname. Detailed configurations are summarized in Table~\ref{tab:hparams}. For the progressive warm-start experiments with existing checkpoints, we utilize JetMoE-8B~\cite{shen2024jetmoereachingllama2performance} and OLMoE-7B~\cite{muennighoff2025olmoeopenmixtureofexpertslanguage} as our base models and convert them into the UE architecture.

\textbf{Scaling Settings and Budget Metrics.}
Our goal is to assess whether \aname provides consistent benefits across both model scale and routing granularity. The two baselines above correspond to fine-grained routing (12 experts with top-$2$) and coarse-grained routing (64 experts with top-$8$), respectively. We study two expansion settings: \textbf{width expansion}, where \aname does not increase activated parameters or total physical parameters and only enlarges the expressivity via Virtual Width; and \textbf{depth expansion}, where \aname performs depth scaling by sharing FFN parameters across depth. We report results under three budgets: \textbf{Act} (activated parameters), \textbf{TP} (total physical parameters), and \textbf{VP} (virtual parameters), where VP accounts for the effective capacity when Virtual Width is considered. This protocol directly tests whether \aname yields a better scaling frontier under both fine-grained and coarse-grained MoE regimes.



\begin{table*}[t]
\centering
\caption{Training and routing hyperparameters used in our experiments. Values are reported for the Qwen-3 style MoE backbones and are kept identical between MoE and \aname unless explicitly noted.}
\vskip 0.15in
\small

\setlength\tabcolsep{4pt}

\begin{tabular*}{\textwidth}{@{\extracolsep{\fill}}lcc@{}}
\toprule

\textbf{Hyperparameter} & \textbf{Value} & \textbf{Notes} \\
\midrule
Sequence length & 4096 & shared \\
Global batch size & 256 & shared \\
Optimizer & AdamW & $\beta=(0.9,0.95)$, weight decay $0.1$ \\
LR schedule & cosine & shared \\
Peak LR & $5\times10^{-4}$ & shared \\
Warmup fraction & $5\%$ tokens & shared \\
MoE aux loss coef. & $10^{-3}$ & MoE baseline \\
DN (data/param ratio) & 200 & shared \\
\midrule
Connectivity group size $G$ & 5 & layers sharing the same expert mask \\
\midrule
MoE-160M: layers & 18 & Qwen-3 style \\
MoE-160M: hidden size & 768 & Qwen-3 style \\
MoE-160M: heads / groups & $12 / 4$ & attention / query groups \\
MoE-160M: experts / top-$k$ & $12 / 2$ & per layer \\
MoE-160M: MoE FFN hidden & 192 & per expert \\
\midrule
MoE-700M: layers & 16 & Qwen-3 style \\
MoE-700M: hidden size & 1536 & Qwen-3 style \\
MoE-700M: heads / groups & $16 / 4$ & attention / query groups \\
MoE-700M: experts / top-$k$ & $64 / 8$ & per layer \\
MoE-700M: MoE FFN hidden & 96 & per expert \\
\midrule
UELB $\alpha_{\text{loc}}$ & 1.0 & calibrated \\
UELB $\alpha_{\text{u}}$ & 0.5 & calibrated \\
UELB warmup $b_0$ & 0.75 & initial universal logit bias \\
UELB warmup $r$ & 5\% & token \% \\
Universal Router $\beta$ & 0.1 & contextual pathway weight \\
Fast-weight lr $\eta$ & 0.1 & online update learning rate \\
\bottomrule
\end{tabular*}

\label{tab:hparams}
\vspace{-2mm}
\end{table*}

\section{Topology Settings and Fair Comparisons}
When comparing connectivity schemes (e.g., all-to-all, fixed window sharing, and staggered rotational sharing), we keep the \textbf{window size} $W$ and the \textbf{number of activatable experts per layer} fixed. The only difference is the connectivity structure (which experts are reachable at each layer). This ensures that topology ablations reflect optimization and routing effects rather than changes in activated-parameter budgets.

\end{document}